\documentclass[runningheads]{llncs}
\usepackage{amsmath}
\usepackage{graphicx}
\usepackage{booktabs}
\usepackage{lscape}
\usepackage{multirow}
\usepackage{float}
\usepackage{placeins}
\usepackage{comment}
\usepackage{subcaption} % Add this in the preamble

\begin{document}
%Template: Springer Lecture Notes in Computer Science (3)
\title{Missing data imputation for noisy time-series data and applications in healthcare}
%Imputation Techniques for Noisy Health Time Series Data
\titlerunning{Missing data imputation for noisy time-series healthcare data}
 \author{Lien P. Le\inst{1,2,4} \and
 Xuan-Hien Nguyen Thi\inst{1,2} \and
 Thu Nguyen\inst{5} \and
 Michael A. Riegler\inst{5} \and
 Pål Halvorsen\inst{5} \and
 Binh T. Nguyen\inst{1,2,3}\thanks{Corresponding Author: Binh T. Nguyen (ngtbinh@hcmus.edu.vn)}}
% %
 \authorrunning{Le et al.}
 \institute{Faculty of Mathematics and Computer Science, University of Science, Ho Chi Minh City, Vietnam 
 \and
 Vietnam National University, Ho Chi Minh City, Vietnam
 %\url{http://www.springer.com/gp/computer-science/lncs} 
 \and
 AISIA Research Laboratory, Ho Chi Minh City, Vietnam
 \and
 University of Medicine and Pharmacy at Ho Chi Minh City, Ho Chi Minh City, Vietnam
 \and
 Simula Metropolitan, Oslo, Norway\\
 }
\maketitle              % typeset the header of the contribution
\begin{abstract}
Healthcare time series data is vital for monitoring patient activity but often contains noise and missing values due to various reasons such as sensor errors or data interruptions. Imputation, i.e., filling in the missing values, is a common way to deal with this issue. In this study, we compare imputation methods, including Multiple Imputation with Random Forest (MICE-RF) and advanced deep learning approaches (SAITS, BRITS, Transformer) for noisy, missing time series data in terms of MAE, F1-score, AUC, and MCC, across missing data rates (10\%-80\%). Our results show that MICE-RF can effectively impute missing data compared to deep learning methods and the improvement in classification of data imputed indicates that imputation can have denoising effects. Therefore, using an imputation algorithm on time series with missing data can, at the same time, offer denoising effects.  

\end{abstract}
%
%
%
%\vspace{-10mm}
\section{Introduction}\label{sec-intro}
%\vspace{-3mm}
In healthcare, time series data is essential for monitoring patient activity, offering insights into movement patterns, daily routines, and overall physical well-being. This data is often captured using wearable devices like actigraphy monitors, which provide minute-by-minute recordings of physical activity \cite{le2023data}. However, such data often includes noise and missing values due to interruptions in data collection, non-compliance, or sensor errors. Effective imputation of missing values is crucial to ensure accurate analysis in clinical studies.

Many missing data handling techniques have been proposed in the research community, from the methods that directly handle missing data \cite{nguyen2022dper,dinh2021clustering}, to imputation techniques \cite{mice} that fill in the missing values. For time series data, a variety of methods have been developed as well \cite{hua2024impact}. Simple methods such as the Last Observation Carried Forward (LOCF) technique and linear interpolation \cite{le2023data} are still frequently used but can introduce bias and inaccuracies.
% , especially when the data is not stationary or when there are sudden shifts in sensor readings. 
Classical techniques such as
K-Nearest Neighbors (KNN), MICE \cite{mice} are still widely used as well.
% and matrix factorization \cite{NIPS2016_85422afb} methods offer alternatives, KNN estimates missing values by averaging the values of the nearest neighbors, while matrix factorization methods decompose the data matrix into lower-rank matrices, using matrix completion to estimate and fill the missing values.  MICE (Multiple Imputation by Chained Equations) \cite{mice} goes further by iteratively generating multiple estimates for missing values, refining them with each iteration to provide a more robust imputation. 
% Additionally, advanced machine learning models like XGBoost \cite{qu2024xgboost} have been applied for time series imputation, leveraging their predictive capabilities to fill for missing values. 
Also, in recent years, many deep learning-based methods have been developed for time series missing data, such as Generative Adversarial Networks (GANs) \cite{Pingi-Longitudinal-2023}, BRITS \cite{cao2018brits}, due to their ability in modeling complex patterns. %but can be computationally intensive.

In this paper, we compare the performance of MICE-RF and state-of-the-art deep learning methods for time series imputation (SAITS \cite{du2023saits}, BRITS\cite{cao2018brits}, and Transformer \cite{yildiz2022multivariate}), across various missing data rates (10\% to 80\%). Note that mean square error is not suitable as an evaluation measure for noisy data due to its squaring effect. Therefore, after using classification algorithms like Logistic Regression, AdaBoost, and KNN, we assessed the imputation quality with metrics such as  MAE, F1-score, AUC, and MCC to assess the imputation quality. Through this study, by incorporating seasoned features of time series, MICE-RF, which is Multiple Imputation by Chained Equations (MICE) using RandomForestRegressor with max\_iter and random\_state set to 5 and 0 respectively (with other hyperparameters as default), effectively utilizes both past and future values for time series imputation. Also, we illustrate that imputation methods can have denoising effects. Therefore, for noisy time series with missing values, imputation not only fills in the missing values but also helps denoise the data.

Our contribution can be summarized as follows: (i) We compare various imputation methods for noisy time series imputation; (ii) We show that MICE-RF can outperform state-of-the-art time series imputation techniques for noisy, missing data; (iii) We draw the connection between using time period in MICE-RF with the predicting strategy in Long Short-Term Memory (LSTM) where the predictions are based on both in the past and future time-steps; (iv) We illustrate that imputation techniques can have denoising effects for noisy time series data.
%\vspace{-5mm}
\section{Methods under Comparision}\label{sec-methods}
%\vspace{-3mm}
\subsection{MICE imputation with Random Forest}
MICE (Multiple Imputation by Chained Equations) is a statistical method used to handle missing data by creating multiple complete datasets through iterative imputation and averaging the results. When Random Forest is used as the predictive model in MICE (\textbf{MICE-RF}), it allows capturing complex relationships between variables, models non-linearities and interactions between features and handles mixed data, which is common in many real-world datasets with missing values. To use MICE-RF for time series data, we can reshape the time series to leverage the temporal structure. 
Suppose that we have a dataset of time series $\left\{\left(x^{(1)}, y^{(1)}\right), \left(x^{(2)}, y^{(2)}\right),\ldots,\left(x^{(n)}, y^{(n)}\right)\right\},$  where 
\begin{equation}
    x^{(i)} =
\begin{pmatrix}
x^{(i)}_1 \\
x^{(i)}_2 \\
\vdots \\
x^{(i)}_{T_i}
\end{pmatrix},
\end{equation}
is the $i^{th}$ time series of the dataset, $x_j^{(i)} \in \mathbf{R}$ denotes the $j^{\text{th}}$ epoch of $x^{(i)}$,
$y^{(i)} \in \mathbf{R}$ is the label of the $i^{th}$ time series, $T_{i}^{(i)}$ is the length of $x^{(i)}$.
To make the imputation for the dataset, we impute in each time series $x^{(i)}$, for $i \in \{1, 2, \dots, n\}$.
% The following will clarify how MICE-RF works in imputing sub-time series $x^{(i)}$
We can reshape each $x^{(i)}$ into matrix $\mathcal{X}^{(i)}$ as the following:%the $k$ x $T_{\text{period}}$ matrix:
\begin{equation}
\renewcommand{\arraystretch}{1.5}
\mathcal{X}^{(i)} = \left( 
\begin{array}{c@{\hspace{0.8cm}}c@{\hspace{0.8cm}}c@{\hspace{0.8cm}}c}
x_1^{(i)} & x_2^{(i)} & \dots & x_{T_{\text{period}}}^{(i)} \\
x_{T_{\text{period}}+1}^{(i)} & x_{T_{\text{period}}+2}^{(i)} & \dots & x_{2T_{\text{period}}}^{(i)} \\
\vdots & \vdots & \ddots & \vdots \\
x_{(k-1)T_{\text{period}}+1}^{(i)} & x_{(k-1)T_{\text{period}}+2}^{(i)} & \dots & x_{T_i}^{(i)} \\
\end{array}
\right),
\label{eq:MICE-RF_matrix}
\end{equation}
where $T_{period}$ is chosen as not only the the period of $x^{(i)}$ but also a divisor of $T_i$ ,  $k=\frac{T_i}{T_{period}}$.
Then, we apply MICE-RF on the resulted matrix $\mathcal{X}^{(i)}$ in (\ref{eq:MICE-RF_matrix}) to impute missing data. After that, we reshape $\mathcal{X}^{(i)}$ back to the original size to have the imputed version of $x^{(i)}$. 
% And so on we have the imputed data for the other sub-time series, this means the imputed dataset of time series.

So, from equation \ref{eq:MICE-RF_matrix}, we can see that utilizing MICE-RF to impute missing data for the time series means we are using the past and future time values to infer the current value. Moreover, reshaping time series like this helps us take advantage of the temporal information in the time series while using MICE with Random Forest.

\subsection{SAITS}
The Self-Attention-Based Imputation for Time Series (SAITS) \cite{du2023saits} is a novel algorithm for imputing missing values in multivariate time series data. It uses self-attention mechanisms to capture complex temporal relationships, with two main components: Diagonally Masked Self-Attention (DMSA) blocks and a Weighted Combination Block.

DMSA blocks are specifically designed to enhance the model's understanding of time dependencies. 
%By masking diagonal entries in the attention matrix, each time step is influenced by other time steps rather than its own information, allowing for a richer representation of temporal patterns. 
The model uses a stack of two DMSA blocks, refining the representation of time-series data with each layer.
Following the DMSA blocks, the Weighted Combination Block combines the outputs of the DMSA layers. It dynamically weighs different learned representations, adapting the model to the data's unique characteristics. This ensures that the model can balance the influence of different time steps based on their relevance to the imputation task.

SAITS employs a joint-optimization strategy with two tasks: the Masked Imputation Task (MIT) and the Observed Reconstruction Task (ORT). In MIT, a portion of the observed values is masked, and the model learns to impute them accurately, training for robust imputation. The ORT, on the other hand, focuses on reconstructing observed values, ensuring that the model retains consistency with the original data. The combined loss from both tasks allows the model to achieve a balance between accurate imputation and data consistency.

% Overall, SAITS efficiently capture the temporal dependencies and feature correlations, making it more effective than traditional recurrent models for time series imputation. The joint-optimization approach further ensures that the model is both accurate and stable in training.
%\vspace{-3mm}
\subsection{BRITS}

Bidirectional Recurrent Imputation for Time Series (BRITS) \cite{cao2018brits} is an approach that leverages bidirectional RNNs for imputing missing data by learning temporal relationships in both forward and backward directions. This dual processing enables BRITS to capture the dynamics of time series data more comprehensively, making it effective for real-world applications without imposing strict assumptions about data generation processes.
% There are four steps that BRITS conclude, that is recurrent dynamics for imputation, imputation as variables, conducting consistency loss And training.
% BRITS represents time series data as a sequence with multiple features across different time steps. A masking vector indicates whether a particular feature at a time step is observed or missing, guiding the model in the imputation process.

%BRITS is based on a bidirectional RNN architecture with LSTM units. The forward LSTM processes the sequence from the beginning to the end, using past observations to estimate missing values. Conversely, the backward LSTM processes the data in reverse, using future observations to make predictions. This dual processing enables BRITS to consider both past and future data points when filling in gaps. Here, the missing values are treated as learnable parameters that are updated during training. The model combines the predictions from both the forward and backward LSTMs to estimate each missing value. %This combined approach improves the accuracy of the imputation by considering multiple perspectives.
BRITS uses a bidirectional RNN with LSTM units, where the forward LSTM processes data from start to end, and the backward LSTM processes it in reverse, allowing both past and future observations to estimate missing values. Missing values are treated as learnable parameters, and updated during training. The model combines predictions from both directions to fill in gaps effectively.

To ensure that forward and backward estimates are aligned, BRITS includes a consistency constraint. This encourages the two sets of predictions to match, leading to more coherent imputations. The training process involves minimizing a loss function that balances the accuracy of the imputations with this consistency constraint. By learning from both past and future observations, BRITS 
%achieves a higher level of imputation accuracy, making 
makes a great solution for handling missing values in time series data.
%\vspace{-5mm}
\subsection{Transformer}
%\vspace{-3mm}
%Transformer for time series missing value imputation \cite{yildiz2022multivariate} is a method that leverages self-attention to capture both local and long-range dependencies within the data. Its self-attention mechanism, which calculates relationships between data points, allows it to excel in modeling both local and long-range dependencies. This makes it particularly suitable for handling time series data with missing values. The core of the Transformer is its self-attention mechanism, which calculates the relationships between all data points in a sequence. This allows the model to weigh the importance of different time steps and focus on the most relevant information. By doing so, the Transformer can understand the dependencies and interactions between time points, even when some values are missing.
The Transformer for time series imputation \cite{yildiz2022multivariate} uses self-attention to capture both local and long-range dependencies within the data, making it suitable for handling time series data with missing values. Its self-attention mechanism assesses relationships between data points, allowing the model to weigh the importance of different time steps, focus on the most relevant information, and understand dependencies across time steps. By doing so, the Transformer can understand the dependencies and interactions between time points, even when some values are missing.

The Transformer generates imputed values by using the contextual information derived from the self-attention layers. Each time step is represented by a context-aware vector, which is used to predict the missing values. The model's ability to combine information from both nearby and distant time steps ensures accurate imputations.

To improve its performance, the model is trained on data with simulated missing patterns. This helps the Transformer gain robustness and generalize better to real-world data, where missing values may occur in unpredictable ways. The result is a model for handling missing data for time series that effectively bridges the gaps in complex sequences.
%\vspace{-3mm}
\section{Experiments}\label{sec-experiment}
%\vspace{-3mm}
%\subsection{Experiment settings}\label{experiment-setting}
\subsection{Datasets}
We evaluate the methods on the following three datasets:
\begin{itemize}
\item  \textbf{Psykose} \cite{Jakobsen2020}: This is a univariate time series data in healthcare for schizophrenia prediction. The dataset comprises actigraph data measuring gravitational acceleration units per minute collected from 22 patients with schizophrenia and 32 healthy controls.
%\item  \textbf{Psykose} \cite{Jakobsen2020}: This is a univariate time series data in healthcare for schizophrenia prediction. The dataset comprises actigraph data measuring gravitational acceleration units per minute collected from 22 patients with schizophrenia, all of whom were hospitalized in a long-term open psychiatric ward at Haukeland University Hospital and were on antipsychotic medication. Additionally, it includes actigraphy data from 32 healthy controls.
% including 20 females and 12 males, consisting of 23 hospital staff, 5 nursing students, and 4 individuals recruited from a general practitioner. None of the controls had a history of psychotic or mood disorders. 
%All participants wore the actigraph devices for an average of 12.7 days in both the patient and control groups.
\item  \textbf{Depresjon} \cite{Garcia:2018:NBP:3083187.3083216}: This is a univariate time series data for depression prediction. The dataset contains actigraphy data from 23 patients with unipolar and bipolar depression (condition group), including 5 who were hospitalized during the data collection period and 18 outpatients. Additionally, it includes actigraphy data from 32 non-depressed individuals (control group).

%\item  \textbf{Depresjon} \cite{Garcia:2018:NBP:3083187.3083216}: This is a univariate time series data in healthcare for depression prediction. The dataset contains actigraphy data from 23 patients with unipolar and bipolar depression (condition group), including 5 who were hospitalized during the data collection period and 18 outpatients. Additionally, it includes actigraphy data from 32 non-depressed individuals (control group).
% , comprised of 23 hospital employees, 5 students, and 4 former patients with no current psychiatric symptoms. 
%For each participant, a CSV file is provided, containing the actigraph data recorded over time. The columns in these files include timestamp (at one-minute intervals), date (of measurement), and activity (activity level recorded by the actigraph device). Some other demography data are also provided including the number of days of measurements.
%\item  \textbf{HTAD} \cite{10.1007/978-3-030-67835-7_17}: The HTAD dataset (Home-Tasks Activities Dataset) is a multi-source and multivariate time series data for home-tasks prediction. It contains wrist-accelerometer and audio data from three individuals performing various household activities, such as sweeping, brushing teeth, washing hands, and watching TV. These activities represent tasks necessary for independent living. The dataset aims to support the development of assistive technologies, particularly in domains like elderly care and mental health monitoring. This paper uses the accelerometer information with motion captured from the x, y, and z axes for imputation.
\item  \textbf{HTAD} \cite{10.1007/978-3-030-67835-7_17}: The HTAD (Home-Tasks Activities Dataset) is a multi-source, multivariate time series dataset for predicting essential home tasks. It includes wrist-accelerometer and audio data from three individuals performing activities like sweeping, brushing teeth, washing hands, and watching TV. This dataset aids in developing assistive technologies, particularly for elderly care and mental health. 
%This study uses accelerometer data from x, y, and z axes for imputation.
\end{itemize}
%\vspace{-7mm}
\subsection{Experimental setup} 
For all the datasets, we first classify using full data and get the metrics as the ground truth. 
% We get the full activity data as the same method in the paper correspondingly. For example, the full data of the Psykose dataset includes the time series activity data of the number of first full days in the file "days.csv" of the dataset. 
For the Psykose and Depresjon datasets, which are univariate time series datasets,
% including the activity daily of participants and have seasonal information, 
with MICE-RF, based on the seasoned characteristic of time series, we reshape the univariate time series to multivariate based on $T_{period}$ values of 1440, 60, 30, and 15 minutes indicating the time of one day and the decreased time. We also use corresponding n\_steps for deep learning methods. After tuning, the best result for MICE-RF is at $T_{period}$ of 15 minutes, and for deep learning, n\_steps of 15 or 30 minutes yield similar results. Thus, we select 15 minutes for both $T_{period}$ and n\_steps.
%With the HTAD dataset, which is a multivariate time series data and there is no seasonal information and it is not possible to find a common divisor among the length of time series in the dataset, we do not reshape the time series for MICE-RF, and we use n\_steps equal to 1 to have an equivalence in choosing parameters.
With the HTAD dataset, which includes multivariate time series performing various household activities without seasonal information and lacks a common divisor across the lengths of time series, we do not reshape the time series for MICE-RF and set n\_steps to 1 to maintain consistency in parameter selection. SAITS and Transformer are evaluated with both the parameter n\_layers equal to 1 and 2. 

For each dataset, we generate missing data with missing rates increasing from 10\% to 80\%, with a spacing of 10\%. 
% We alternately make the imputation for these missing datasets by four imputation methods MICE-RF, BRITS, Transformer, and SAITS. 
For evaluation, we calculate the Mean Absolute Error (MAE) between the imputed and original data and use classification accuracy metrics such as F1-score, Area Under the Curve (AUC), and Matthews Correlation Coefficient (MCC). For each of the datasets in the experiments, we use one of the classification techniques utilized as the baseline evaluation according to the comparison of the original dataset papers \cite{Jakobsen2020,Garcia:2018:NBP:3083187.3083216,10.1007/978-3-030-67835-7_17}. Therefore, the classification methods utilized in Psykose, Depresjon, and HTAD datasets are Logistic Regression \cite{Jakobsen2020}, Ada Boost \cite{Garcia:2018:NBP:3083187.3083216} and KNN \cite{10.1007/978-3-030-67835-7_17}, respectively.
The code is operated in Python with the sklearn package for MICE-RF and the pypots package for the deep learning methods.
\subsection{Experiment results}\label{Imputation-performance}
\begin{figure}[h!]
    \centering
    \begin{subfigure}[b]{0.5\linewidth}
        \centering
        \includegraphics[width=\linewidth]{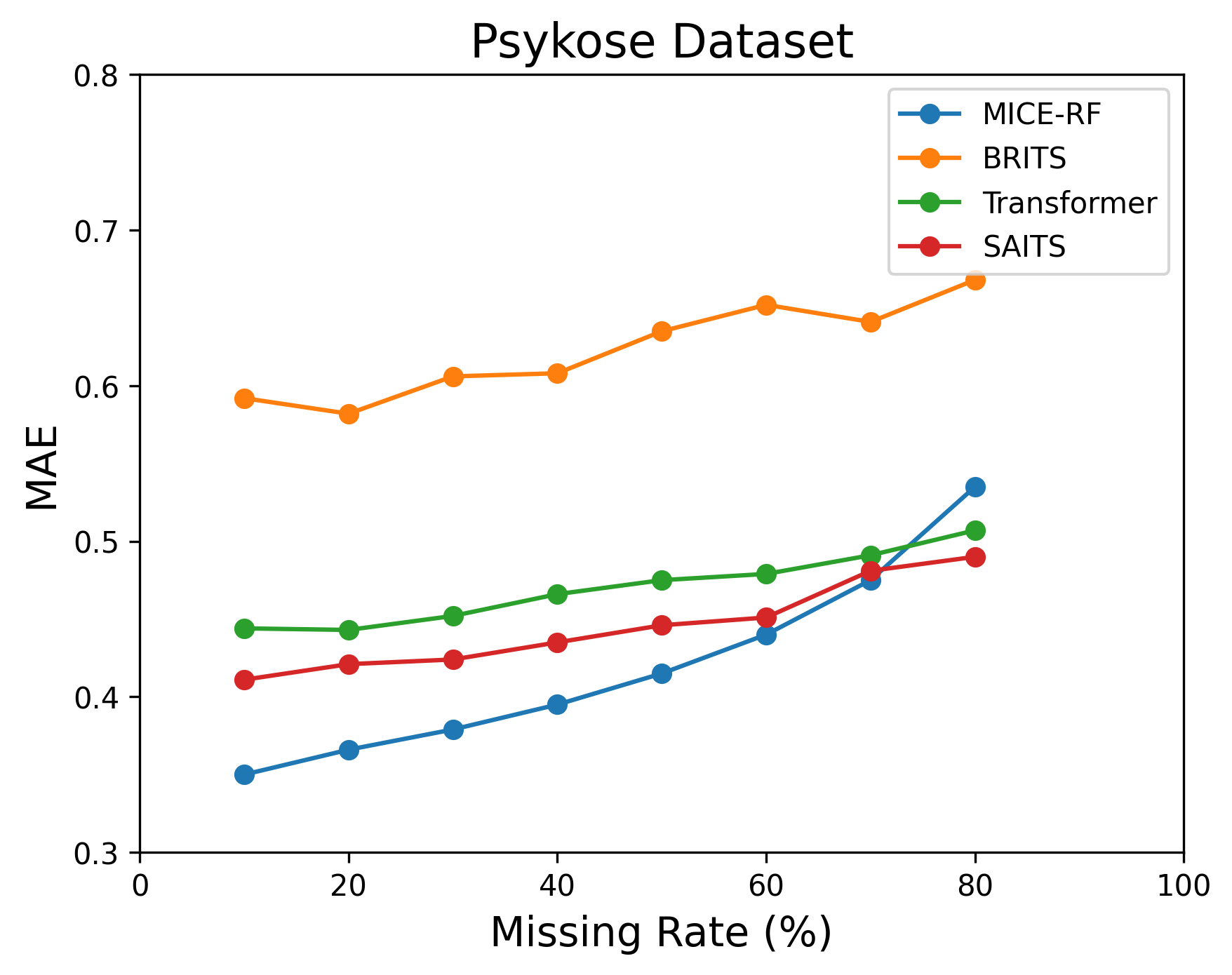}
        \caption{n\_layers=1}
        \label{fig:mae_psykose-1nlayer}
    \end{subfigure}
    \hspace{-2mm}
    \begin{subfigure}[b]{0.5\linewidth}
        \centering
        \includegraphics[width=\linewidth]{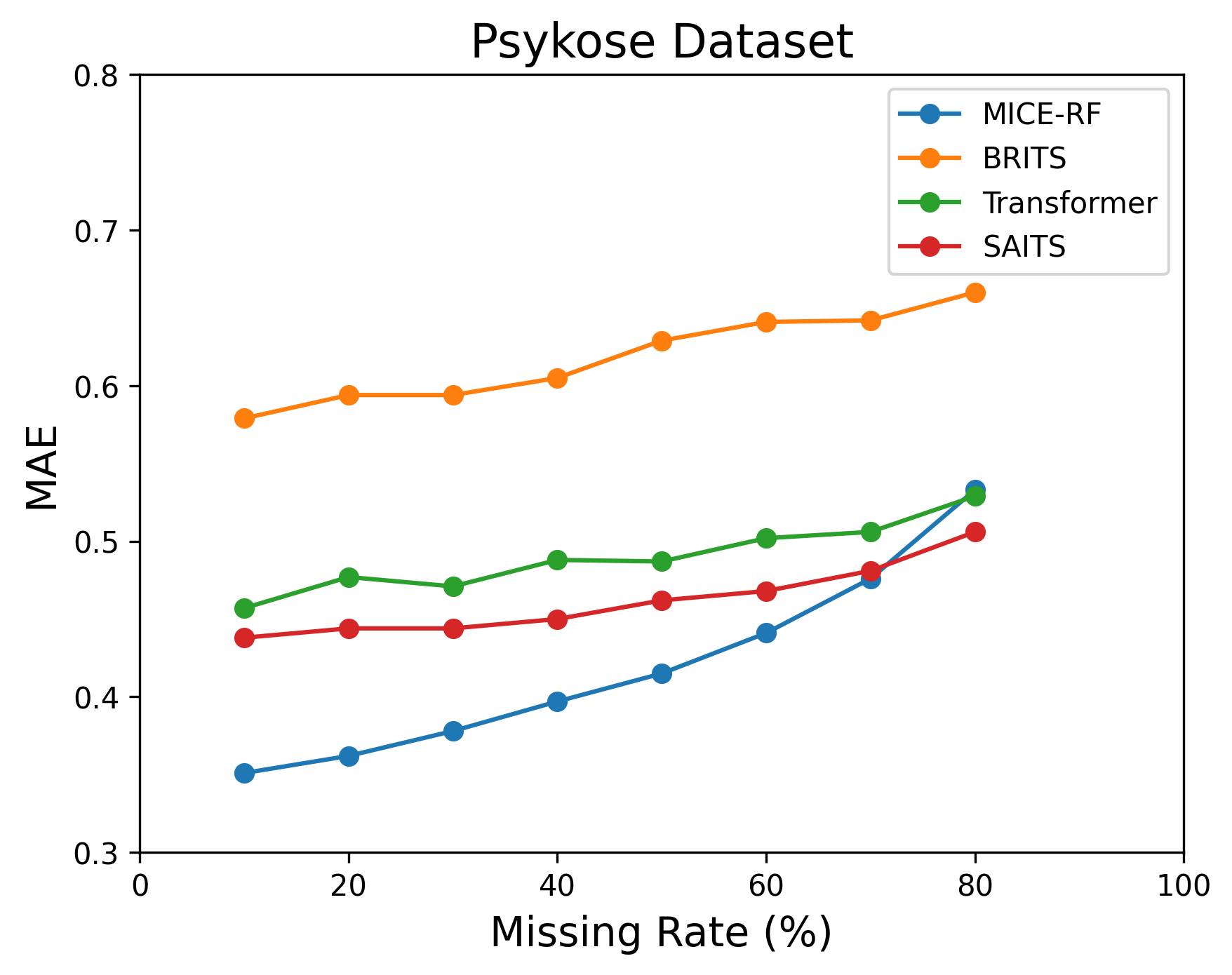}
        \caption{n\_layers=2}
        \label{fig:mae_psykose-2nlayer}
    \end{subfigure}
    \caption{The performance of methods on Psykose Dataset for different layers}
    \label{fig:mae_psykose-comparison}
\end{figure}

\begin{figure}[h!]
    \centering
    \begin{subfigure}[b]{0.5\linewidth}
        \centering
        \includegraphics[width=\linewidth]{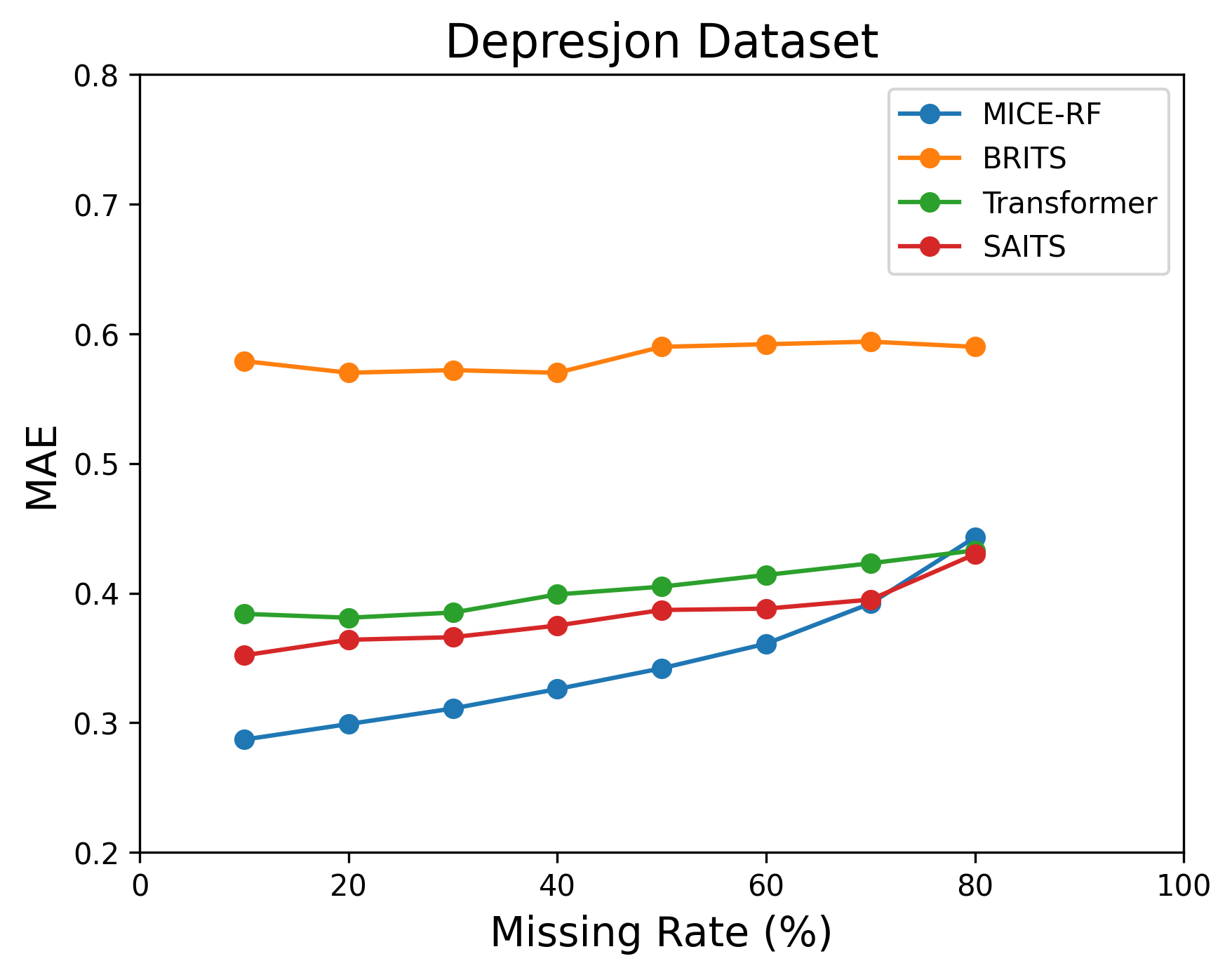}
        \caption{n\_layers=1}
        \label{fig:mae_depresjon-1nlayer}
    \end{subfigure}
    \hspace{-2mm}
    \begin{subfigure}[b]{0.5\linewidth}
        \centering
        \includegraphics[width=\linewidth]{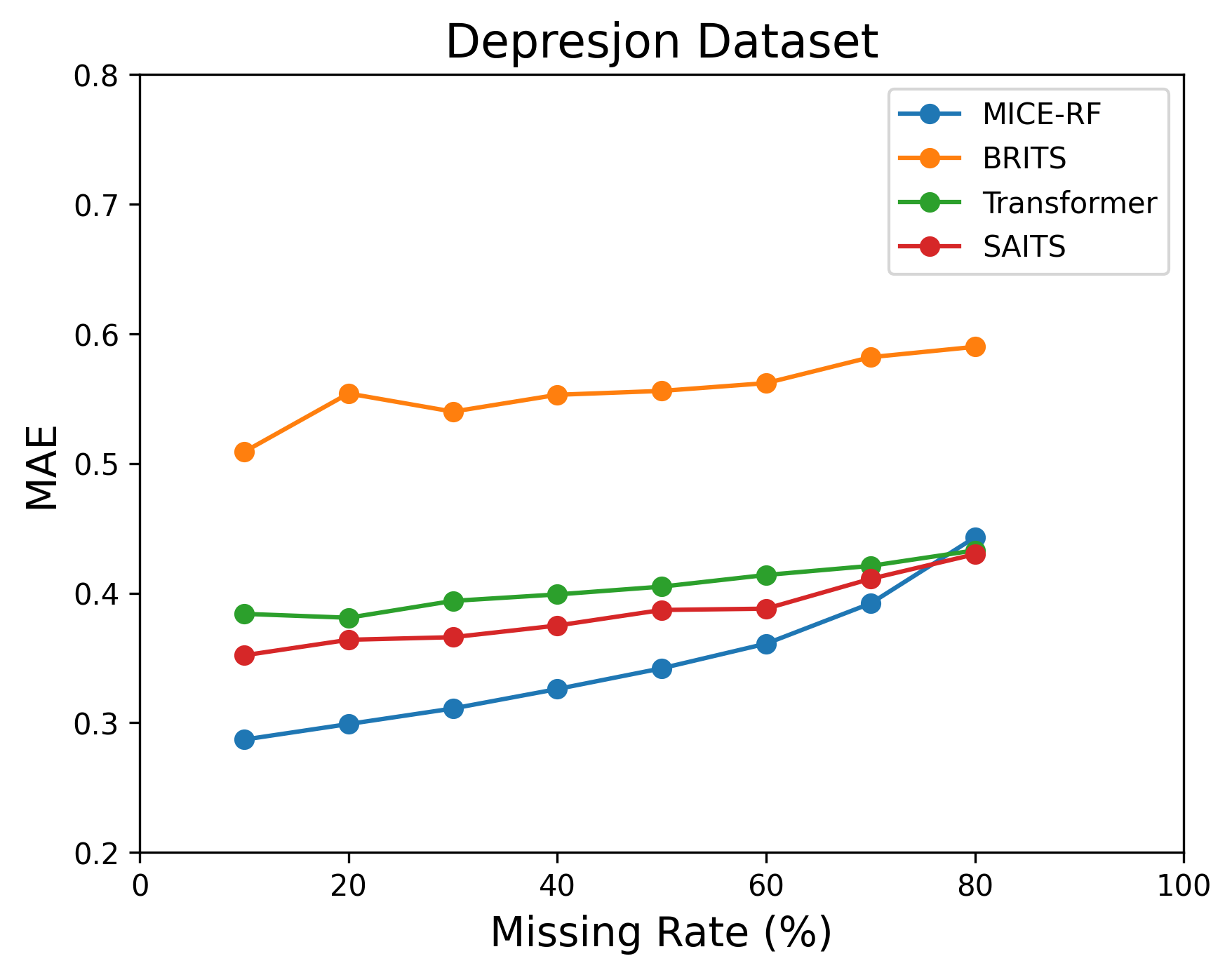}
        \caption{n\_layers=2}
        \label{fig:mae_depresjon-2nlayer}
    \end{subfigure}
    \caption{The performance of methods on Depresjon Dataset for different layers}
    \label{fig:mae_depresjon-comparison}
\end{figure}

For both the Psykose dataset and the Depresjon dataset (Figure \ref{fig:mae_psykose-1nlayer}, Figure \ref{fig:mae_psykose-2nlayer}, Figure \ref{fig:mae_depresjon-1nlayer}, and Figure \ref{fig:mae_depresjon-2nlayer}), with missing rates under 60\%,  MICE-RF shows the best overall imputation accuracy, followed by SAITS, Transformer, and BRITS, for both the experiments with the parameter n\_layer equal to 1 and 2. However, for missing rates greater than 60\%, MICE-RF begins to increase MAE compared with the other methods. Additionally, deep learning methods with n\_layer = 1 yield lower MAE than those with n\_layer = 2.
%Plot for MAE comparison 
%Psykose
% \begin{figure}[h!]
%     \centering
%     \includegraphics[width=0.9\linewidth]{psykose_mae_results_plot_1layer.png}
%     \caption{Performance of methods (n\_layers=1) on Psykose Dataset}
%     \label{fig:mae_psykose-1nlayer}
% \end{figure}
% \begin{figure}[h!]
%     \centering
%     \includegraphics[width=0.9\linewidth]{imputation_mae_results_psykose_2layers.png}
%     \caption{Performance of methods (n\_layers=2) on Psykose Dataset}
%     \label{fig:mae_psykose-2nlayer}
% \end{figure}
%Depresjon
%MAE of HTAD
\begin{figure}[h!]
    \centering
    \begin{subfigure}[b]{0.5\linewidth}
        \centering
        \includegraphics[width=\linewidth]{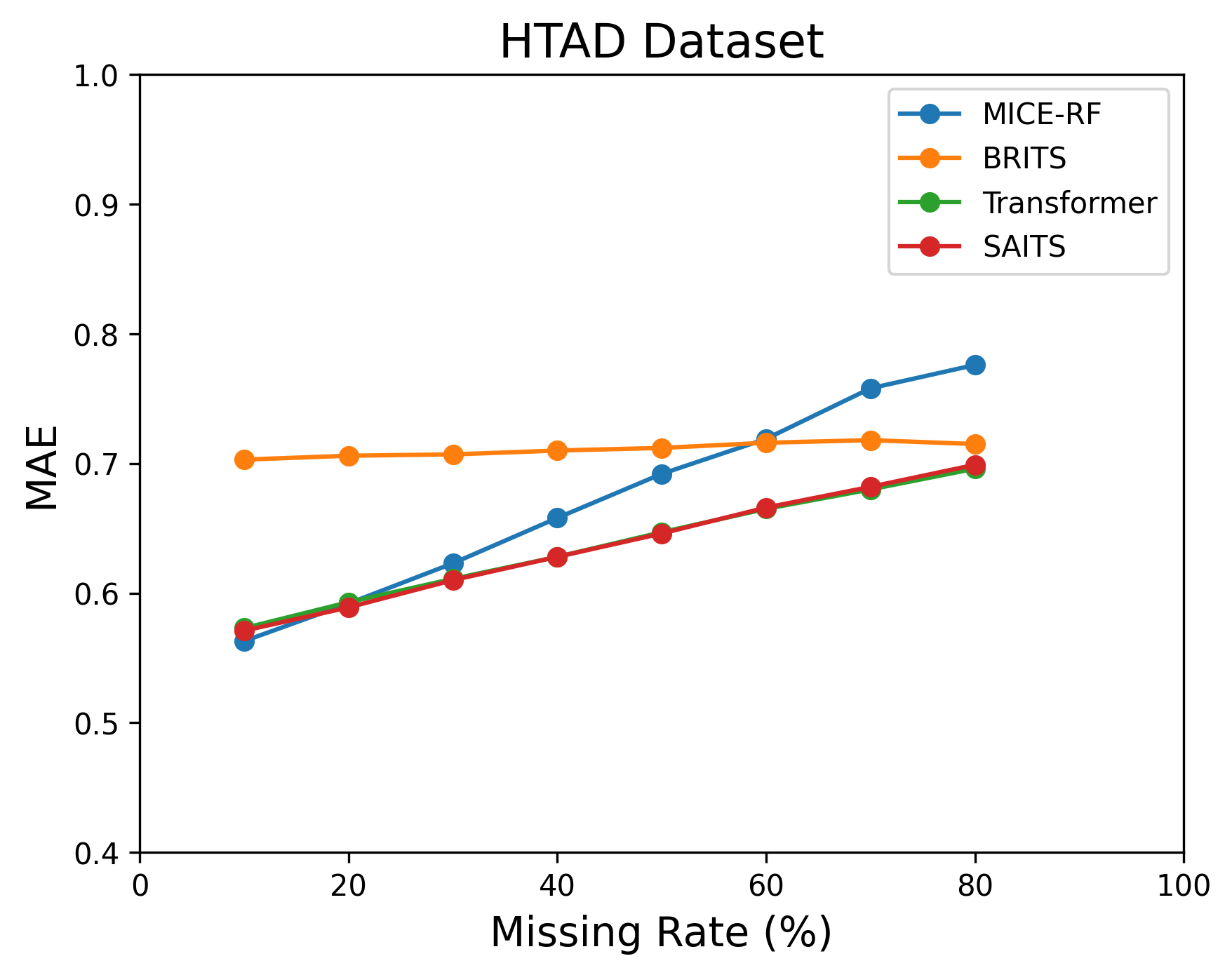}
        \caption{n\_layers=1}
        \label{fig:mae_htad-1nlayerl}
    \end{subfigure}
    \hspace{-2mm}
    \begin{subfigure}[b]{0.5\linewidth}
        \centering
        \includegraphics[width=\linewidth]{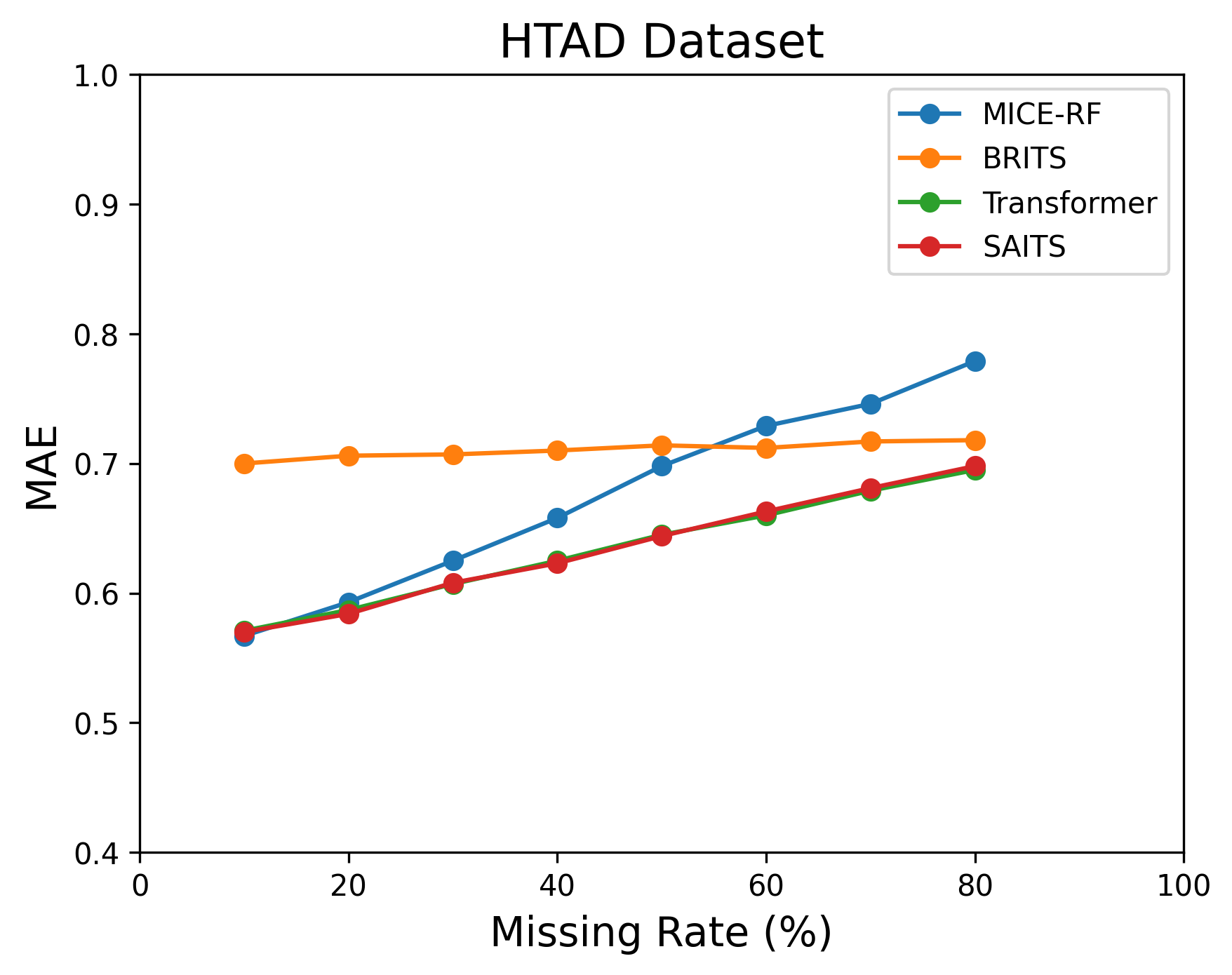}
        \caption{n\_layers=2}
        \label{fig:mae_htad-2nlayer}
    \end{subfigure}
    \caption{The performance of methods on HTAD Dataset for different layers}
    \label{fig:mae_htad-2nlayer}
\end{figure}
% \begin{figure}[h!]
%     \centering
%     \includegraphics[width=0.9\linewidth]{imputation_mae_HTAD_1layer.png}
%     \caption{Performance of methods (n\_layers=1) on HTAD Dataset}
%     \label{fig:mae_htad-1nlayerl}
% \end{figure}
% \begin{figure}
%     \centering
%     \includegraphics[width=0.9\linewidth]{imputation_mae_htad_2layer.png}
%     \caption{Performance of methods (n\_layers=2) on HTAD Dataset}
%     \label{fig:mae_htad-2nlayer}
% \end{figure}
For the HTAD dataset (Figure \ref{fig:mae_htad-1nlayerl}, Figure \ref{fig:mae_htad-2nlayer}), at 10\% missing rate, all methods show fairly similar performance. The MAE remains low, with SAITS and MICE-RF showing a slight edge over the others. As the missing rate increases, MICE-RF shows the most noticeable rise in MAE, implying that its performance degrades with more missing data. SAITS consistently maintains a lower error, indicating its robustness.
% \subsection{Downstream classification task}\label{downstream_classification}

% Next, as mentioned in section \ref{experiment-setting}, we make classifications on the full original data of three datasets and get three metrics (F1-score, AUC, MCC) as the baseline for the corresponding classification of imputed data. 
Next, regarding classification results, from Tables \ref{table:classify-psykose-1layer}, \ref{table:classify-psykose-2layer}, \ref{table:classify-depresjon-1layer}, \ref{table:classify-depresjon-2layer}, \ref{table:classify-htad-1layer}, and \ref{table:classify-htad-2layer}, one can notice that most of metrics are either greater than or not too low compared with ground truth. For the comparison of methods, with the Psykose dataset and the HTAD dataset, MICE-RF outperformed the remaining methods. However, BRITS and Transformer perform the improvement in the imputation of the Depresjon dataset. Interestingly, MICE-RF is not good in terms of MAE for the HTAD dataset but has the best result in terms of downstream tasks for this time series. This illustrates that MAE may not be a good measure of imputation quality for noisy data. 

Interestingly, from Table \ref{table:classify-psykose-1layer}, we can see that the baseline, which is the results of classification on the original data (before missing data simulation), is 0.848. Meanwhile, the results on imputed data at a missing rate of 10\%-30\% by MICE-MF is higher than that. Imputed data using BRITS and SAITS sometimes also gives higher accuracy, although less occasionally than MICE-RF. Similar things can be observed for AUC and MCC on this dataset and across the other tables. So, in general, the values of metrics go up when the missing rate increases (and sometimes it may decrease after certain points). This is most evident in the case of the classification of the HTAD dataset. This illustrates that imputation can also denoise the data, in addition to filling in the missing entries, and this also has positive effects on imbalanced data, as F1-score and MCC give insights into the performance of imbalanced data.
% (the improvements in the F1-score or MCC can reveal the capability of improvement in the classification of imbalanced data).
%\vspace{-5mm}
\section{Discussion}\label{sec-discuss} 
%\vspace{-3mm}
Random Forests, by design, are less prone to overfitting due to their ensemble nature, especially when the number of trees is large. In the context of MICE-RF, the multiple imputation process adds further robustness, as the model is iteratively refined, and multiple imputations account for uncertainty. Deep learning models are highly flexible and powerful, but this flexibility comes with an increased risk of overfitting, particularly when data is scarce or has noisy patterns. Regularization techniques and careful tuning are required to prevent overfitting in SAITS and other deep-learning models.
%Random Forests are less prone to overfitting due to their ensemble nature, especially with a large number of trees. In MICE-RF, multiple imputations enhance robustness by iteratively refining the model and accounting for uncertainty. Deep learning models, while powerful, are more prone to overfitting, particularly with scarce or noisy data. Regularization and careful tuning are essential to prevent overfitting in SAITS and similar deep-learning models.
% It is important to note that MICE + RF can be computationally intensive and may require substantially more time for imputation compared to other methods, especially as the dataset size or complexity increases. This trade-off between accuracy and computational cost should be considered depending on the context of the application. 
However, MICE-RF also has its own limitations. Specifically, Random Forest is computationally expensive, especially when applied in an iterative manner as in MICE. For long time series with many missing values, the imputation process can become slow. 
% If too many lagged variables are used as predictors, or if the lags span too far into the past, the Random Forest model may overfit the imputed values, making them overly smooth or too closely tied to recent trends. Selecting the right number of lags is crucial.  

The performance of these imputation methods is highly dependent on several parameters. For MICE-RF, the number of iterations and the choice of period $T_{period}$ have a significant impact on both the accuracy and the computational time required. The smaller $T_{period}$ is, the more computational time is required as more sub-series are created. 
%However, if $T_{period}$, RF can overfit, leading to lower accuracy.
Next, note that SAITS, BRITS, and Transformer, being neural network-based methods, are sensitive to hyperparameters. As mentioned in section \ref{Imputation-performance}, imputation results with n\_layer = 1 and n\_layer = 2 have a difference. Moreover, n\_steps is also an important value to the performance of methods in terms of not only the accuracy of the imputation but also computing time. %Similar to the choice of the parameter $T_{period}$ of MICE-RF, we do not always find a common divisor between the lengths of the time series as the value of n\_steps.
%\vspace{-5mm}
 % Random Forests handle both continuous and categorical data seamlessly. This is useful for time series datasets that might have mixed data types, such as financial data, where both numerical and categorical features are common. MICE-RF can impute missing values for any type of data without needing separate models or preprocessing steps. Deep learning models are typically designed to work with continuous, real-valued data. Categorical variables often need to be one-hot encoded or embedded, which adds complexity to the model and can lead to suboptimal performance if not handled properly.
 % \subsection{The dependence of imputation performance on the parameters}
 \section{Conclusion}
%\vspace{-3mm}
By evaluating MICE-RF, BRITS, Transformer, and SAITS on three noisy healthcare time series datasets with different characteristics,
% (two univariate time series data and one multivariate time series data), 
it is evident that MICE-RF consistently gives the best MAE in two univariate time series datasets for missing rates under 60\%. This highlights its effectiveness as a robust tool for univariate time series imputation, especially when adapted to leverage temporal dependencies through seasonal information.
%showcasing that it can be a powerful tool for univariate time series imputation when adapted to account for temporal dependencies using seasonal information. 
MICE-RF offers flexibility in capturing complex, non-linear relationships and interactions between observations. However, care must be taken to balance the computational cost and handle large gaps in missing data. 
% with the data reshaped based on the seasonal information. 

In the case of the multivariate time series dataset HTAD, which does not contain the periodicity, when the data is not reshaped, deep learning methods such as SAITS, BRITS, and Transformer give better MAE than MICE-RF. 

Additionally, all evaluated imputation methods have been effective in ensuring stability and even improving downstream classification with all time series data containing missing values evaluated due to noise. This demonstrates that these imputation methods can simultaneously impute and denoise time series data. 
Additionally, the method with the lowest MAE does not always yield the best classification outcomes.
% It may depend on the pattern of missing data.
In summary, some imputation methods not only impute missing data but also help reduce noise when dealing with noisy, incomplete healthcare time series data.
% In the future, we plan to develop a denoising algorithm for time series based on imputation.
\vspace{-5mm}
\begin{landscape}
\begin{table}[h!]
    \centering
    \caption{Results of the downstream classification task on the Psykose dataset (n\_layers =1). The higher, the better. Values in bold are the best in each metric.}
    \setlength{\tabcolsep}{3pt} % Adjust column spacing (default is 6pt)
\renewcommand{\arraystretch}{1.25} % Adjust row height (1.0 is the default)
\begin{tabular}{|c|cccc|cccc|cccc|}
\hline
    \multirow{2}{*}{M.R}. & \multicolumn{4}{c|}{F1-score (baseline: 0.848)} & \multicolumn{4}{c|}{AUC (baseline: 0.904)} & \multicolumn{4}{c|}{MCC (baseline: 0.687)} \\
    %\midrule
    \cline{2-13}
  & MICE-RF &  BRITS & Transformer &  SAITS & MICE-RF &  BRITS & Transformer &  SAITS & MICE-RF &  BRITS & Transformer &  SAITS \\
  \hline
10 & \textbf{0.853} & \textbf{0.853} & 0.838 & \textbf{0.853} & 0.905 & 0.901 & 0.901 & \textbf{0.906} & \textbf{0.698} & \textbf{0.698} & 0.667 & 0.696 \\
20 & \textbf{0.863} & 0.848 & 0.848 & 0.838 & \textbf{0.907} & 0.899 & 0.899 & 0.902 & \textbf{0.718} & 0.687 & 0.688 & 0.668 \\
30 & \textbf{0.858} & 0.848 & 0.838 & 0.848 & 0.904 & 0.895 & 0.900 & \textbf{0.910} & \textbf{0.707} & 0.692 & 0.668 & 0.688 \\
40 & \textbf{0.848} & 0.824 & 0.843 & 0.833 & \textbf{0.903} & 0.893 & 0.902 & 0.896 & \textbf{0.688} & 0.637 & 0.679 & 0.661 \\
50 & \textbf{0.843} & 0.824 & 0.814 & 0.838 & \textbf{0.904} & 0.891 & 0.884 & 0.901 & \textbf{0.676} & 0.648 & 0.619 & 0.668 \\
60 & \textbf{0.838} & \textbf{0.838} & 0.828 & 0.833 &\textbf{ 0.903} & 0.892 & 0.885 & 0.892 & \textbf{0.670} & 0.666 & 0.650 & 0.659 \\
70 & \textbf{0.848} & 0.828 & 0.838 & 0.784 & 0.898 & \textbf{0.908} & 0.893 & 0.862 & \textbf{0.688} & 0.648 & 0.672 & 0.565 \\
80 & \textbf{0.843} & 0.819 & 0.814 & 0.824 & \textbf{0.891} & 0.876 & 0.864 & 0.889 & \textbf{0.676} & 0.634 & 0.621 & 0.639 \\
\bottomrule
\end{tabular}
\label{table:classify-psykose-1layer}
\vspace{0.5cm}
%\end{table}

%\FloatBarrier

%\begin{table}[H]
    \centering
    
    \caption{Results of the downstream classification task on the Psykose dataset (n\_layer =2). The higher, the better. Values in bold are the best in each metric}
    \setlength{\tabcolsep}{3pt} % Adjust column spacing (default is 6pt)
\renewcommand{\arraystretch}{1.25} % Adjust row height (1.0 is the default)
\begin{tabular}{|c|cccc|cccc|cccc|}
%\toprule
\hline
\multirow{2}{*}{M.R}. & \multicolumn{4}{c|}{F1-score (baseline: 0.848)} & \multicolumn{4}{c|}{AUC (baseline: 0.904)} & \multicolumn{4}{c|}{MCC (baseline: 0.687)} \\
    %\midrule
    \cline{2-13}
  & MICE-RF &  BRITS & Transformer &  SAITS & MICE-RF &  BRITS & Transformer &  SAITS & MICE-RF &  BRITS & Transformer &  SAITS \\ 
  \hline
         10 &  0.843 &     \textbf{0.848} &           0.838 &     \textbf{0.848} &   \textbf{0.903} &      0.897 &            0.899 &      0.901 &   0.676 &      \textbf{0.688} &            0.668 &      \textbf{0.688} \\
               20 &  \textbf{0.853} &     0.848 &           0.843 &     0.838 &   0.904 &      0.903 &            \textbf{0.907} &      0.899 &   \textbf{0.698} &      0.688 &            0.677 &      0.667 \\
               30 &  \textbf{0.853} &     0.848 &           0.848 &     0.843 &   \textbf{0.907} &      0.894 &            0.899 &      0.899 &   \textbf{0.698} &      0.690 &            0.688 &      0.677 \\
               40 &  \textbf{0.863} &     0.833 &           0.838 &     0.824 &   \textbf{0.905} &      0.894 &            0.892 &      0.885 &   \textbf{0.718} &      0.659 &            0.667 &      0.641 \\
               50 &  \textbf{0.853} &     0.804 &           0.838 &     0.828 &   \textbf{0.907} &      0.889 &            0.884 &      0.886 &   \textbf{0.698} &      0.603 &            0.668 &      0.652 \\
               60 &  \textbf{0.853 }&     0.843 &           0.848 &     0.824 &   \textbf{0.905} &      0.892 &            0.901 &      0.885 &   \textbf{0.698} &      0.679 &            0.690 &      0.639 \\
               70 &  \textbf{0.858} &     0.833 &           0.833 &     0.838 &   \textbf{0.896} &      0.894 &            0.882 &      0.890 &   \textbf{0.708} &      0.659 &            0.659 &      0.672 \\
               80 &  0.838 &     \textbf{0.848} &           0.760 &     0.814 &   \textbf{0.899} &      0.888 &            0.867 &      0.892 &   0.666 &      \textbf{0.694} &            0.512 &      0.628 \\
\hline
\end{tabular}
\label{table:classify-psykose-2layer}
\end{table}
\end{landscape}

\begin{landscape}
    \begin{table}[h!]
         \centering
    \caption{Results of the downstream classification task on the Depresjon dataset (n\_layer =1). The higher, the better. Values in bold are the best in each metric}
    \setlength{\tabcolsep}{3pt} % Adjust column spacing (default is 6pt)
\renewcommand{\arraystretch}{1.25} % Adjust row height (1.0 is the default)
\begin{tabular}{|c|cccc|cccc|cccc|}
\hline
    \multirow{2}{*}{M.R}. & \multicolumn{4}{c|}{F1-score (baseline: 0.731)} & \multicolumn{4}{c|}{AUC (baseline: 0.794)} & \multicolumn{4}{c|}{MCC (baseline: 0.396)} \\
    %\midrule
    \cline{2-13}
  & MICE-RF &  BRITS & Transformer &  SAITS & MICE-RF &  BRITS & Transformer &  SAITS & MICE-RF &  BRITS & Transformer &  SAITS \\
  \hline
         10 &  0.741 &     0.731 &           \textbf{0.744} &     0.725 &   0.808 &      \textbf{0.819} &            0.818 &      0.811 &   0.413 &      0.394 &            \textbf{0.426} &      0.369 \\
               20 &  0.735 &     \textbf{0.738} &           0.712 &     0.728 &   0.816 &      \textbf{0.823} &            0.784 &      0.815 &   0.391 &      \textbf{0.413} &            0.350 &      0.384 \\
               30 &  0.741 &     0.748 &           0.735 &     0.741 &   \textbf{0.786} &      \textbf{0.836} &            0.800 &      0.784 &   0.417 &      \textbf{0.425} &            0.400 &      0.419 \\
               40 &  0.718 &     0.738 &           \textbf{0.757} &     0.748 &   0.794 &      0.823 &            \textbf{0.841} &      0.817 &   0.374 &      0.405 &            \textbf{0.441} &      0.426 \\
               50 &  0.706 &     0.738 &           \textbf{0.748} &     0.744 &   0.769 &      \textbf{0.822} &            0.801 &      0.813 &   0.333 &      0.395 &            \textbf{0.422} &      0.412 \\
               60 &  0.693 &     0.751 &           \textbf{0.783} &     0.741 &   0.759 &      0.829 &            \textbf{0.851} &      0.833 &   0.293 &      0.435 &            \textbf{0.507} &      0.403 \\
               70 &  0.696 &     \textbf{0.786} &           0.735 &     0.744 &   0.757 &      \textbf{0.840} &            0.784 &      0.830 &   0.320 &      \textbf{0.516} &            0.409 &      0.418 \\
               80 &  0.735 &     \textbf{0.761} &           0.748 &     0.735 &   0.792 &     \textbf{ 0.824} &            0.814 &      0.812 &   0.409 &      \textbf{0.453} &            0.432 &      0.391 \\
\bottomrule
\end{tabular}
\label{table:classify-depresjon-1layer}
%\end{table}
\vspace{0.5cm}

%\begin{table}[H]
         \centering
    \caption{Results of the downstream classification task on the Depresjon dataset (n\_layer =2). The higher, the better. Values in bold are the best in each metric}
    \setlength{\tabcolsep}{3pt} % Adjust column spacing (default is 6pt)
\renewcommand{\arraystretch}{1.25} % Adjust row height (1.0 is the default)
\begin{tabular}{|c|cccc|cccc|cccc|}
\hline
    \multirow{2}{*}{M.R}. & \multicolumn{4}{c|}{F1-score (baseline: 0.731)} & \multicolumn{4}{c|}{AUC (baseline: 0.794)} & \multicolumn{4}{c|}{MCC (baseline: 0.396)} \\
    %\midrule
    \cline{2-13}
  & MICE-RF &  BRITS & Transformer &  SAITS & MICE-RF &  BRITS & Transformer &  SAITS & MICE-RF &  BRITS & Transformer &  SAITS \\
  \hline
    10 & \textbf{ 0.748} &     0.722 &           0.725 &     0.735 &  \textbf{ 0.812} &      0.808 &            0.789 &      0.805 &   \textbf{0.420} &      0.367 &            0.386 &      0.407 \\
               20 &  \textbf{0.744} &     \textbf{0.744} &           0.738 &     0.731 &   0.807 &      0.809 &           \textbf{ 0.816} &      0.807 &   0.424 &      \textbf{0.432} &            0.413 &      0.396 \\
               30 &  0.725 &     \textbf{0.761} &           0.748 &     0.718 &   0.783 &      \textbf{0.847} &            0.841 &      0.809 &   0.375 &      \textbf{0.461} &            0.428 &      0.354 \\
               40 &  0.715 &     \textbf{0.757} &           0.725 &     0.741 &   0.777 &      \textbf{0.829} &            0.810 &      0.800 &   0.347 &      \textbf{0.443} &            0.366 &      0.417 \\
               50 &  \textbf{0.754} &     0.748 &           0.731 &     0.715 &   0.804 &      0.811 &            \textbf{0.824} &      0.788 &   \textbf{0.452} &      0.436 &            0.386 &      0.343 \\
               60 &  0.735 &     0.735 &           0.751 &     \textbf{0.780} &   0.794 &      \textbf{0.831} &            0.816 &      0.812 &   0.412 &      0.395 &            0.431 &      \textbf{0.505} \\
               70 &  0.706 &     \textit{0.780} &           0.770 &     0.751 &   0.776 &      \textbf{0.852} &            0.826 &      0.822 &   0.340 &      \textbf{0.502} &            0.492 &      0.449 \\
               80 &  0.702 &     \textbf{0.770} &           0.744 &     0.735 &   0.776 &      \textbf{0.832} &            0.823 &      0.826 &   0.327 &      \textbf{0.482} &            0.416 &      0.400 \\
\bottomrule
%\hline
\end{tabular}
\label{table:classify-depresjon-2layer}
\end{table}
\end{landscape}

\begin{landscape}

\vspace{0.5cm}

\begin{table}[]
\centering
\setlength{\tabcolsep}{3pt} % Adjust column spacing (default is 6pt)
\renewcommand{\arraystretch}{1.25} % Adjust row height (1.0 is the default)
    \caption{Results of the downstream classification task on the HTAD dataset (n\_layer=1). The higher, the better. Values in bold are the best in each metric}
    \begin{tabular}{|c|cccc|cccc|cccc|}
    %\toprule
    \hline
    \multirow{2}{*}{M.R}. & \multicolumn{4}{c|}{F1-score (baseline: 0.843)} & \multicolumn{4}{c|}{AUC (baseline: 0.934)} & \multicolumn{4}{c|}{MCC (baseline: 0.816)} \\
    %\midrule
    \cline{2-13} 
    %\hline
           & MICE-RF & BRITS & Transformer & SAITS & MICE-RF & BRITS & Transformer & SAITS & MICE-RF & BRITS & Transformer & SAITS \\
    %\midrule
    \hline

 10.0 &    \textbf{0.859} &      \textbf{0.859} &       0.855 &      0.849 &   0.955 &      \textbf{0.961} &       0.957 &      0.954 &   \textbf{0.836} &       0.835 &        0.83 &       0.823 \\
 20.0 &    0.886 &      0.863 &       \textbf{0.887} &      0.876 &   \textbf{0.968} &      0.961 &       0.966 &      \textbf{0.968} &   0.867 &        0.84 &       \textbf{0.868} &       0.855 \\
 30.0 &    \textbf{0.909} &      0.869 &       0.896 &      0.898 &   \textbf{0.976} &      0.963 &       0.967 &      0.969 &   \textbf{0.894} &       0.846 &       0.878 &       0.881 \\
 40.0 &    \textbf{0.931} &      0.882 &       0.914 &      0.916 &    \textbf{0.980} &      0.968 &       0.977 &       \textbf{0.980} &   \textbf{0.919} &       0.862 &       0.899 &       0.902 \\
 50.0 &    \textbf{0.943} &      0.902 &       0.929 &      0.932 &   \textbf{0.984} &      0.973 &       0.982 &      0.983 &   \textbf{0.933} &       0.886 &       0.918 &        0.920 \\
 60.0 &     \textbf{0.960} &      0.913 &       0.947 &       0.950 &   0.988 &      0.977 &       0.986 &      \textbf{0.989} &   \textbf{0.953} &       0.899 &       0.938 &       0.942 \\
 70.0 &    \textbf{0.971} &       0.940 &       0.958 &      0.966 &   \textbf{0.992} &      0.984 &        0.990 &       0.99 &   \textbf{0.966} &        0.930 &       0.951 &        0.960 \\
 80.0 &    \textbf{0.977} &       0.960 &       0.964 &      0.972 &   \textbf{0.992} &       0.990 &        0.909 &      0.991 &   \textbf{0.974} &       0.953 &       0.958 &       0.967 \\
\bottomrule
\end{tabular}
\label{table:classify-htad-1layer}

\vspace{0.5cm}
    \centering
    \setlength{\tabcolsep}{3pt} % Adjust column spacing (default is 6pt)
\renewcommand{\arraystretch}{1.25} % Adjust row height (1.0 is the default)
    \caption{Results of the downstream classification task on the HTAD dataset (n\_layer=2). The higher, the better. Values in bold are the best in each metric}
    \begin{tabular}{|c|cccc|cccc|cccc|}
    %\toprule
    \hline
    \multirow{2}{*}{M.R}. & \multicolumn{4}{c|}{F1-score (baseline: 0.843)} & \multicolumn{4}{c|}{AUC (baseline: 0.934)} & \multicolumn{4}{c|}{MCC (baseline: 0.816)} \\
    %\midrule
    \cline{2-13} 
    %\hline
           & MICE-RF & BRITS & Transformer & SAITS & MICE-RF & BRITS & Transformer & SAITS & MICE-RF & BRITS & Transformer & SAITS \\
    %\midrule
    \hline
         10 & \textbf{0.860}  & 0.853 & 0.857 & 0.859 & \textbf{0.961} & 0.957 & 0.958 & 0.959 & \textbf{0.836} & 0.828 & 0.833 & 0.835 \\
         20 & 0.879  & 0.863 & 0.870 & \textbf{0.887} & \textbf{0.965} & 0.960 & 0.962 & \textbf{0.965} & 0.858 & 0.840 & 0.848 & \textbf{0.867} \\
         30 & \textbf{0.903} & 0.871 & 0.893 & \textbf{0.903} & 0.970 & 0.965 & 0.970 & \textbf{0.974} & \textbf{0.887} & 0.849 & 0.876 & 0.886 \\
         40 & \textbf{0.937} & 0.881 & 0.913 & 0.923 & \textbf{0.982} & 0.968 & 0.977 & 0.976 & \textbf{0.926} & 0.860 & 0.898 & 0.910 \\
         50 & \textbf{0.946} & 0.889 & 0.928 & 0.932 & 0.983 & 0.971 & 0.982 & \textbf{0.984} & \textbf{0.936} & 0.870 & 0.915 & 0.921 \\
         60 & \textbf{0.966} & 0.924 & 0.950 & 0.957 & \textbf{0.990} & 0.978 & 0.987 & 0.988 & \textbf{0.961} & 0.911 & 0.941 & 0.950 \\
         70 & \textbf{0.969} & 0.938 & 0.953 & 0.963 & \textbf{0.989} & 0.984 & \textbf{0.989} & \textbf{0.989} & \textbf{0.964} & 0.928 & 0.945 & 0.957 \\
         80 & \textbf{0.974} & 0.957 & 0.966 & 0.969 & \textbf{0.992} & 0.989 & 0.990 & 0.990 & \textbf{0.970} & 0.949 & 0.960 & 0.963 \\
    %\bottomrule
    \hline
    \end{tabular}
    \label{table:classify-htad-2layer}
   
\end{table}
\end{landscape}
\bibliographystyle{splncs04}
\bibliography{ref}
\end{document}